\pdfoutput=1

\documentclass[11pt]{article}

\usepackage[final]{coling}
\usepackage{times}
\usepackage{latexsym}
\usepackage{xcolor}
\usepackage[T1]{fontenc}

\usepackage[utf8]{inputenc}

\usepackage{microtype}

\usepackage{inconsolata}

\usepackage{graphicx}
\usepackage{arabtex,utf8}
\usepackage{comment,pbox,multirow,adjustbox}
\usepackage{pifont,graphics,textcomp,array,float,caption,subcaption,todo}
\newcommand\CS{CSW}

\title{
A Survey of Code-switched Arabic NLP: \\Progress, Challenges, and Future Directions
}

  \author{Injy Hamed,$^\lambda$ Caroline Sabty,$^\xi$ Slim Abdennadher,$^\xi$ \\\textbf{Ngoc Thang Vu,$^\sigma$ Thamar Solorio,$^\lambda$$^\delta$ Nizar Habash$^\lambda$$^\mu$}\\
$^\lambda$MBZUAI \quad 
$^\xi$German International University, Egypt \quad
$^\sigma$University of Stuttgart \\
$^\delta$University of Houston \quad 
$^\mu$New York University Abu Dhabi\\
  {\tt \{injy.hamed,thamar.solorio\}@mbzuai.ac.ae} \\
  {\tt \{caroline.sabty,slim.abdennadher\}@giu-uni.de}\\ 
  {\tt  thang.vu@ims.uni-stuttgart.de ~ nizar.habash@nyu.edu}\\
  }

\begin{document}
\maketitle
\begin{abstract}
Language in the Arab world presents a complex diglossic and multilingual setting, involving the use of Modern Standard Arabic, various dialects and sub-dialects, as well as multiple European languages. This diverse linguistic landscape has given rise to code-switching, both within Arabic varieties and between Arabic and foreign languages. The widespread occurrence of code-switching across the region makes it vital to address these linguistic needs when developing language technologies. In this paper, we provide a review of the current literature in the field of code-switched Arabic NLP, offering a broad perspective on ongoing efforts, challenges, research gaps, and recommendations for future research directions.
\end{list} 


\end{abstract}

\setcode{utf8}
\section{Introduction}
Code-switching (CSW), the act of alternating between multiple languages in the same discourse, is a common linguistic phenomenon in multilingual societies, including Arab countries. With CSW's global prevalence, it has become essential to equip language technologies to effectively handle it to build inclusive and user-friendly tools that cater to the needs of multilingual communities. While there exist several survey papers on CSW \cite{CV16,sitaram2019survey,dougruoz2021survey,winata2023decades}, this paper narrows the focus to the Arabic language, offering more in-depth knowledge and insights for this specific language setup. 
We provide this review, discussing current literature, challenges, and research gaps, with the aim of guiding future research and accelerating progress in this area. 

The paper is organized as follows: \S\ref{sec:language_landscape} provides an overview on the linguistic landscape in the Arab world and historical factors giving rise to CSW; 
\S\ref{sec:cs_types} discusses the types of CSW; \S\ref{sec:annotation_process} describes the paper categorization process; 
\S\ref{sec:overall_statistics} presents overall statistics on the current literature in CSW Arabic NLP; \S\ref{sec:insights_language_resource} and \S\ref{sec:insights_NLP} give further insights into the efforts on data collection, modeling, and guidelines and annotation tools that are useful in the context of CSW; and \S\ref{sec:future_directions} suggests future research directions.


\begin{table*}[t]
    \centering
    \setlength{\tabcolsep}{2pt}
    \small
    \begin{tabular}{|l|l|}
    \hline
    \multicolumn{1}{|c|}{\textbf{\CS~ Type}} & \multicolumn{1}{c|}{\textbf{Example}}\\\hline
        \textbf{Inter-sentential}  &  \multicolumn{1}{r|}{I need to prepare for it.   
        \<أنا عندي إمتحان قريب. >}\\
                                       & \textit{I have an exam coming up.} I need to prepare for it. \\\hline
        \textbf{Extra-sentential}  &
                                        \multicolumn{1}{r|}{ 
                                        \<.>
                                        \<هاشتغل على المشروع بكرا > 
                                        Okay }\\
                                        & Okay \textit{I'll work on the project tomorrow.}\\\hline           
         \textbf{Intra-sentential}  & \multicolumn{1}{r|}{
                                            is in the field of computer vision
                                            \<المشروع بتاعي >
                                            }\\
                                        & \textit{My project} is in the field of computer vision.\\\hline

          \textbf{Morphological \CS}      & \multicolumn{1}{r|}{    
                                             before the deadline
                                             \<النهارده>\
                                             models+\<ال> 
                                             evaluate+\<لازم ن> 
                                            }\\
                                        & \textit{We have to} evaluate  \textit{the} models \textit{today} before the deadline.\\\hline

    \end{tabular}
    \caption{Examples of different CSW types followed by their English translation. The originally Arabic phrases are italicized in the English translation.
    }
    \label{fig:CSWtypes}
\end{table*}

\section{Language in the Arab World}
\label{sec:language_landscape}
Historical events have long influenced the patterns of CSW between Arabic and other languages. During the 7th and 8th centuries, the Islamic conquests brought Arabic into extensive contact with several languages, including Persian \cite{khan2011semitic} and Spanish \cite{thomas2019language}, leading to profound linguistic exchanges. 
Later, from the 16th until the early 20th century, Turkish and Arabic intermingled during the Ottoman Empire, leaving lasting imprints on both languages. 
Afterwards, during the 19th and 20th centuries, several Arab countries were impacted by the spread of European colonialism across the region. 
According to \citet{Dur08} and \citet{CRS+14}, colonization played a main role in shaping the language in the region, where the local languages were overlain by European languages. 
These events and cultural interactions led to linguistic exchanges, giving rise to CSW between Arabic and numerous languages. In contemporary times, globalization and international businesses and education have further intensified this phenomenon, with Arabic speakers increasingly using English and other languages in social and professional settings. 


Besides code-switching to foreign languages, with Arabic being a diglossic language \citep{Fer59}, Arabs also switch between formal Arabic and dialects. The formal Arabic variant, Modern Standard Arabic (MSA), serves as a lingua franca across Arab countries and is typically used in formal contexts. 
Dialects (and sub-dialects), belonging to each country, are used in everyday conversations and informal writings. 
This gives rise to two main types of CSW in the region: (1)~CSW between MSA and dialects, and (2)~CSW between Arabic and foreign languages. Following \citet{ABD+18}, we refer to the former type as \textit{diglossic code-switching } and the latter as \textit{bilingual code-switching}. Bilingual CSW is seen across the region, including Arabic-English in 
Egypt \citep{Abu91}, Jordan \citep{MA94}, 
Palestine \cite{mkahal2016codeswitching}, Saudi Arabia \citep{OI18}, and UAE \citep{Khu03}. A high level of multilingualism, with the mixing of Arabic, English, and French, is found in Morocco \citep{samih2016arabic}, Algeria \citep{baya2016sociolinguistic}, Lebanon \citep{BB11}, and Tunisia \citep{Bao09}.

While Arabic CSW occurs with a wide range of languages, we restrict the scope of this paper to those so far addressed within the field of computational linguistics. We also restrict our scope to CSW language alternation, excluding studies looking into the origin of words in languages influenced by others, such as the investigation presented in \citet{micallef2024cross} for Maltese.

\section{Types of Code-switching}
\label{sec:cs_types}

According to \citet{Pop80}, there are three main types of \CS: inter-sentential, extra-sentential, and intra-sentential. In the case of morphologically rich languages, morphological code-switching \citep{stefanich2019morphophonology} also occurs. 
We elaborate on each type below, providing examples in Table~\ref{fig:CSWtypes}:
\begin{itemize}
    \item \textbf{Inter-sentential \CS} involves switching of languages on the sentence-level. 
    \item \textbf{Extra-sentential \CS} (or \textbf{tag-switching}) 
    involves using tag elements from another language 
    such as fillers, interjections, tags, and idiomatic expressions. 
    \item \textbf{Intra-sentential \CS} 
    involves word-level switching, where \CS~segments must conform to the 
    syntactic rules of both languages. 
    \item \textbf{Morphological \CS} (or \textbf{intra-word \CS}) 
    involves switching on the morpheme-level. Given that Arabic is a morphologically rich language, Arabic-speakers attach Arabic clitics and affixes to foreign words.
\end{itemize}

Another type of language alternation is referred to as \textbf{borrowing}, where loanwords are embedded into sentences without the need for grammatical considerations. According to \citet{Pop80}, borrowing is not considered as a type of CSW. 
\citet{myers1997duelling}, however, considers any embedded segment from the secondary language 
as a case of CSW. Given that 
automatically identifying whether an embedded word is a case of borrowing or CSW is a challenging task, 
the majority of the surveyed efforts in this paper  
consider any presence of foreign tokens as CSW. 
Within the scope of the paper, CSW is used to refer to all the mentioned types of language alternation, including borrowing.

\begin{table*}[t]
    \centering
    \setlength{\tabcolsep}{4pt}
    \small
    \begin{tabular}{l p{13cm}}
    \hline
     \textbf{Category} & \textbf{Options} \\\hline
     \textbf{Language Pairs}&MSA-DA, MSA-Foreign, DA-Foreign, Arabic-Foreign, MSA-DA-Foreign\\ \hline
     \textbf{Venues} & Conference, Workshop, Symposium, Book, Journal, ArXiv, Thesis \\\hline
     \textbf{Methods} & Rule/Linguistic Constraint, Statistical Model, Neural Network, Pre-trained Model\\\hline
     \textbf{NLP Tasks} & \textbf{Text:} 
     Word-level Language Identification, Sentence-level Language Identification, CSW Point Prediction, Dialectness Level Estimation, Dependency Parsing, Fake News Detection, Humor Detection, Humor Generation, Emotion Detection, Language Modeling, Lemmatization, Machine Translation, Micro-Dialect Identification, Named Entity Recognition, Natural Language Entailment, Natural Language Understanding, Part-of-Speech Tagging, Question Answering, Sarcasm Detection, Sentiment Analysis, Semantic Parsing, Spelling Correction and Text Normalization, Summarization, CSW Text Generation, Tokenization, Topic Modeling, Transliteration, Word Analogy, Abusive Language Detection\\
    & \textbf{Speech:} Word-level Language Identification, Sentence-level Language Identification, Automatic Speech Recognition, Speech Synthesis, Speech Translation, Sentence Boundary Detection, Text-to-Speech\\\hline
    \end{tabular}
    \caption{The categories used in the annotation process.}
    \label{table:annotation_process}
\end{table*}
\section{Paper Categorization Process}
\label{sec:annotation_process}
We conduct our search for relevant papers on Google Scholar using keywords that involve \textit{code-switch}, \textit{code-mix}, and \textit{Arabic}. 
Afterwards, we categorize the collected papers, where we take inspiration from the guidelines presented in \citet{winata2023decades}. 
We annotate the papers for multiple categories, as shown in Table~\ref{table:annotation_process}, including the year, venue, and language pairs. For the language pairs, we use five categories specifying CSW between MSA, dialectal Arabic (DA), and foreign language(s): MSA-DA, MSA-Foreign, DA-Foreign, Arabic-Foreign, MSA-DA-Foreign. `Arabic-Foreign' is used when the variant of Arabic (MSA or DA) is not explicitly stated. 
Empirical papers are annotated for the NLP tasks and methodological approaches. Resource papers are annotation for the domain of collected data and the NLP tasks that they support.

\section{Overview on Literature Review}
\label{sec:overall_statistics}
\subsection{Number of Papers Across the Years}
\begin{figure}[h]
    \centering
    \includegraphics[width=0.47\textwidth]{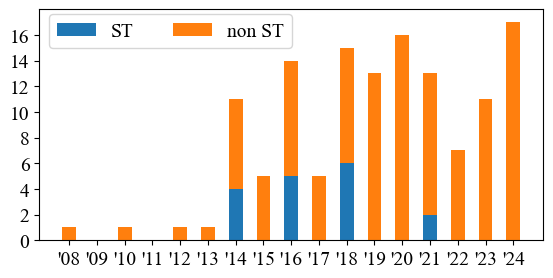}
    \caption{Number of Shared Task (ST) and non-Shared Task (non-ST) papers.}
    \label{fig:overview_years}

\end{figure}

While CSW has been thoroughly studied by linguists since 1980s \citep{Pop80}, it started receiving considerable attention from the Arabic NLP community in 2014. The work on CSW has been greatly motivated by the \textit{Workshop on Computational Approaches to Linguistic Code-Switching}, including shared tasks (ST) which took place in the years of 2014, 2016, 2018, and 2021, reflected in the peaks shown in Figure~\ref{fig:overview_years}. In general, on average, there are 12 papers per year since 2014, reflecting the need for increased attention in this area.
\subsection{Distribution of Papers Across Venue Types}
The majority of the papers are published at conferences (50\%), followed by workshops (36\%, with 13\% absolute belonging to shared tasks), followed by journals (7\%), arXiv papers (4\%) that have not yet been published in venues, and theses (3\%).

\subsection{Evolution in Methodology Methods}
\begin{figure}[h]
    \centering
    \includegraphics[width=\columnwidth]{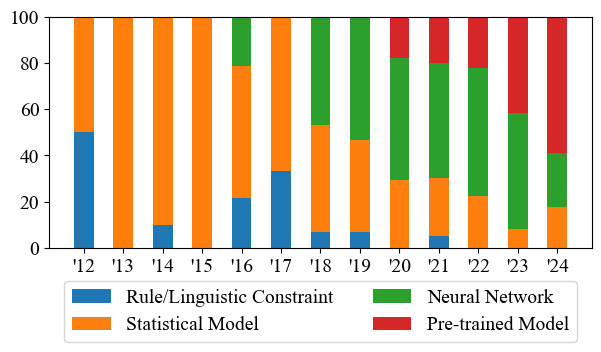}
    \caption{Distribution of papers based on the methods.}
    \label{fig:overview_methods_years}
\end{figure}

We show the evolution in methodology methods in CSW Arabic NLP by plotting the distribution of methods across the years in Figure \ref{fig:overview_methods_years}. 
Comparing these timelines to those presented in \citet{winata2023decades} for CSW overall work, we observe a 3-4 year delay in adopting new methodologies. This is observed for neural-based approaches, which are applied in CSW Arabic NLP in 2016, three years after their utilization in overall work on CSW. Similarly, the use of pretrained models in CSW Arabic NLP started in 2020, compared to 2016 for CSW overall. Despite this delay, we observe alignment with the global NLP trends, showing interest in the Arabic community in advancing CSW research. 


\subsection{Investigated Language Pairs}
\begin{figure*}[t!]
    \begin{subfigure}{0.15\textwidth}
        \includegraphics[height=4.8cm]{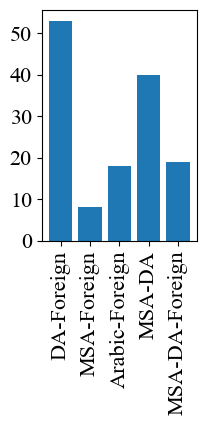}
    \end{subfigure}%
    ~ 
    \begin{subfigure}{0.65\textwidth}
        \includegraphics[height=4.8cm]{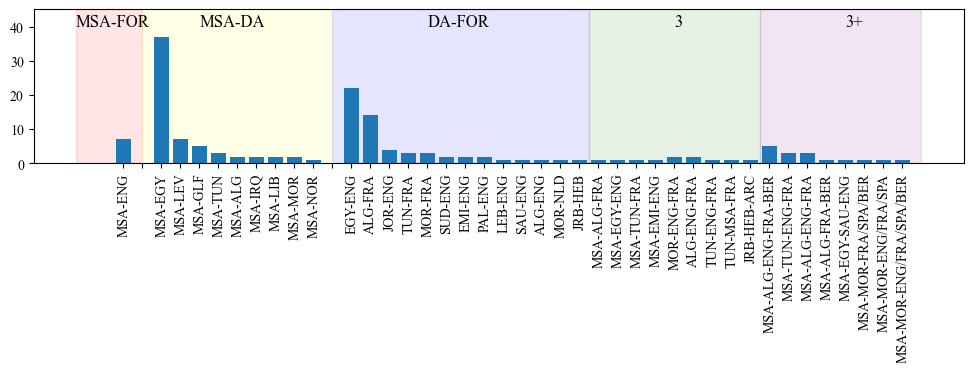}
    \end{subfigure}
    \caption{The number of papers covering the different CSW language setups. We present the distribution as per the annotation guidelines (left) with further language specification (right). Language codes are provided in Appendix~\ref{sec:appendix_lang_codes}.}
    \label{fig:language_pairs}
\end{figure*}

We analyze the papers by language pair categories following our annotation guidelines, specifying Arabic variants and foreign languages when indicated. Figure~\ref{fig:language_pairs} shows these distributions.
We observe that current research is primarily focused on CSW DA-foreign and MSA-DA, which is inline with their prevalent use in real-life. 
For diglossic CSW, MSA-Egyptian is the most studied language pair, while Egyptian-English and Algerian-French are leading for bilingual CSW. Several studies cover more than three languages, however, their count might be exaggerated, as some papers lack sentence-level CSW statistics, making it infeasible to understand the extent of mixing.

\section{Insights on Language Resources}
\label{sec:insights_language_resource}
\paragraph{Dataset Sources} We report the following distribution of datasets' sources: \textit{social media} (36), \textit{transcriptions} (21), \textit{speech recordings} (18), \textit{internet forms and blogs} (11), \textit{news} (8), dialogue (6), 
\textit{songs} (3), \textit{government documents} (2), and \textit{books} (1). 
The domination of social media is expected as it offers a non-expensive source of text, where CSW is likely to occur. However, it is to be noted that CSW phenomena occurring in text could be more restricted than in natural speech \cite{hamed2022arzenST}. For example, in the case of Arabic bilingual CSW, given that Arabic and foreign languages use different scripts, users may be dissuaded to code-switch, especially in the case of intra-word CSW. 
Users may opt for romanization, known as Arabizi \cite{BSM+14}, 
where Arabic words are represented by Latin letters and numerals. 
However, users have different attitudes and preferences regarding the use of Arabizi \cite{alsulami2019sociolinguistic}, where initiatives, such as \textit{BilArabi} (meaning ``\textit{in Arabic}''), have even been established to promote the use of MSA over social media as part of efforts in preserving the Arabic language \cite{taha2019arabic}. 

\paragraph{NLP Tasks' Coverage} In Table \ref{table:nlpTasks_coverage}, we present statistics on the NLP tasks supported by the collected datasets. The complete list of papers is provided in Appendix~\ref{sec:appendix_resource_papers}. We note that the task of language modeling is supported by all collected textual corpora. Otherwise, only \textit{word-level LID} contains a reasonable number of corpora (22). 
This is followed by \textit{ASR} (17),  \textit{MT} (7), \textit{transliteration} (7), and \textit{sentiment analysis} (6). The remaining tasks are not well supported, 
highlighting a significant gap in resources. Further efforts are needed to 
extend existing datasets, offering greater diversity in represented sources, dialects, and tasks.

\begin{table}[t]
\centering
\resizebox{0.96\columnwidth}{!}{
\setlength{\tabcolsep}{2pt}
\begin{tabular}{lrr}
\hline
\multicolumn{1}{c}{\textbf{Task}} & \textbf{E} & \textbf{R}  \\\hline
Text: Word-level Language Identification (LID) & 32  & 22  \\
Text: Named Entity Recognition (NER)& 15 &  3 \\
Text: Machine Translation (MT) & 13 &  7 \\
Text: Sentiment Analysis&  10 &  6 \\
Text: Part-of-Speech Tagging& 6  &  3 \\
Text: Sentence-level Language Identification & 5  & 3  \\
Text: Transliteration& 5  &  7 \\
Text: Language Modeling& 4  & 49  \\
Text: Abusive Language Detection& 2  &  3 \\
Text: Sentence-level Micro-Dialect Identification& 2  &  2 \\
Text: Spelling Correction and Text Normalization&  2 & 1  \\
Text: Dependency Parsing& 2  &  1 \\
Text: Tokenization& 1 &   3\\
Text: Fake News Detection& 1  & 1  \\
Text: Word Analogy& 1  &  1 \\ 
Text: Topic Modeling &	1	&1\\
Text: Question Answering& 1 &   0\\
Text: Lemmatization& 0 &   1\\
Text: Dialectness Level Estimation& 0 & 2\\
Text: Emotion Detection & 0 & 1\\\hline
Speech: Automatic Speech Recognition (ASR)& 22  &  17 \\
Speech: Speech Translation& 2  & 1  \\
Speech: Word-level Language Identification&  1 &  4 \\
Speech: Sentence Boundary Detection& 1  & 11 \\ 
Speech: Sentence-level Language Identification&  0 &  1 \\\hline
\end{tabular}
}
\caption{The coverage of NLP tasks across empirical (E) and resource (R) papers. 
}
\label{table:nlpTasks_coverage}
\end{table}

\section{Insights on NLP Tasks}
\label{sec:insights_NLP}
In Table \ref{table:nlpTasks_coverage}, we present the distribution of empirical papers across tasks (complete list in Appendix~\ref{sec:appendix_empirical_papers}). The most studied tasks are \textit{word-level LID in text}, \textit{ASR}, \textit{NER}, and \textit{MT}. We believe the distribution is greatly affected by Shared Tasks \cite{SBM+14,MAG+16,AAS+18b,chen2021calcs} and PhD theses \cite{elfardy2017perspective,samih2017dialectal,Ama19,adouane2020natural,alghamdi2020towards,sabty2024computational,hamed2024neural}. In this Section, we provide insights into research efforts and challenges in a selection of the NLP tasks.
\subsection{Language Identification (LID)}
\paragraph{Word-level LID}
Word-level LID is a sequence-to-sequence classification task, where the goal is to assign a language label to each word. Only few researchers have tackled this task in the speech domain, covering CSW Algerian-French \cite{AAL17} and MSA-DA \cite{chowdhury2020effects}. The remaining efforts were conducted in the text domain, with social media platforms being the prominent source of text. In the text-domain, the majority of empirical papers covered MSA-DA (19/32), followed by romanized Arabic-foreign (9/32) and Arabic-Arabized foreign (6/32) CSW.

One challenge in this task is the presence of words sharing lexical forms across languages, that are cognates/faux-amis, having similar/different meanings. 
This issue is more pronounced in diglossic CSW due to shared vocabulary between MSA and DA. This introduces ambiguity, especially with the lack of acoustic cues, challenging automatic as well as manual annotations. 
These challenges were discussed in \citet{SBM+14} and \citet{AKS20},
where MSA-DA was found to be the hardest language pair across the explored languages, including Nepali-English, Mandarin-English, Spanish English, and Hindi-English. 
Another challenge that arises, especially in the case of bilingual CSW, is distinguishing between borrowings and CSW. Given the complexity of such a distinction, many researchers do not address this, 
where the decision is left to the annotators’ judgment. 
Alternatively, a language label for borrowed words can be used as proposed in \citet{AD17}. 
Morphological CSW also adds another layer of complexity. While most researchers have handled it using a `mixed' language label, 
\citet{sabty2021language} further expand this label specifying the languages on the morpheme-level in morphological CSW words.
\paragraph{Sentence-level LID} 
For sentence-level LID, we discuss two tasks. The first involves identifying whether a sentence includes CSW, which was  explored for MSA-DA \cite{Alt20} and romanized Arabic-foreign \cite{shehadi2022identifying}.

The second task is \textit{dialect identification}, which involves identifying the Arabic variant of a sentence. This task has been well studied, where it has mostly been defined as a single-label classification task \cite{Bouamor:2018:madar,abdelali2021qadi,abdul2023nadi}. Lately, \citet{abdul2024nadi} modified the task definition into multi-label classification, overcoming the limitations of single labels, including sentences being acceptable under several Arabic variants \cite{keleg2023arabic}. In the scope of CSW, the task has been investigated in the contexts of diglossic \cite{AED15,ERA18} and bilingual \cite{abdul2020toward,Aba18} CSW, where challenges are introduced due to the presence of foreign words and the vocabulary overlap between MSA and DA. 
We note, however, that the current task definition falls short in handling CSW, and we recommend refining it to identify mixed variants within a sentence.

\subsection{Dialectness Level Identification} 
This task has been defined as both a regression \cite{keleg2023aldi} and classification \cite{abdul2024nadi} problem. However, both efforts do not take CSW into account. 
Efforts involving Arabic CSW are limited to guidelines' development and data annotation. 
\citet{badawi1973mustawayat} defines five levels for Egyptian Arabic based on the use of MSA, dialect, as well as foreign languages, reflecting on the educational levels of the participants of each group. \citet{HRD+08} accounts for diglossic CSW when defining the following five levels of dialectness: perfect MSA, imperfect MSA, MSA-dialectal CSW, dialect with MSA incursions, and pure dialect. 
The guidelines were adopted by \citet{hamed2024zaebuc}, 
where transcriptions having a mixture of MSA, Gulf and Egyptian Arabic, and English were annotated accordingly. 

\subsection{Transliteration}
Transliterating romanized Arabic words into Arabic script is a challenging task as romanization is non-standard and unlike HSB transliteration scheme \cite{habash2007arabic}, does not have a direct one-to-one mapping to Arabic words. When compounded with CSW, another layer of complexity is added as foreign words should remain unchanged. Previous approaches \cite{darwish2014arabizi,EAH+14} involved a two-step process of Arabic/foreign word-level LID followed by script conversion. 
\citet{shazal2020unified} presented the first effort for a unified model for both; Arabizi detection and transliteration. 
In terms of the explored language pairs, the main focus of research has been on romanized Arabic-English, where we still lack research on Arabic-Arabized foreign. Little exploration has also been conducted on transliterating Judeo-Arabic with code-switched Hebrew and Aramaic into Arabic \cite{mitelman2024code}. 

\subsection{Machine Translation (MT)}
Despite CSW Arabic MT being only addressed in 13 papers, efforts cover a broad range of dialects and translation directions. 
\citet{chen2021calcs} introduced a shared task, 
covering two-way translation between CSW MSA-Egyptian and English. Research on translating from CSW Arabic-foreign to the foreign 
language has been conducted for several language pairs, covering MSA-English, Egyptian-English, Jordanian-English, Palestinian-English, Algerian-French, Moroccan-French, and Tunisian-French, as outlined in Appendix \ref{sec:appendix_empirical_papers}. 
Less research has focused on translating CSW Arabic-foreign to 
Arabic as well as both directions \cite{hamed2022arzenST,heakl2024arzen}. Translating to the primary language is particularly useful, as it could help bilingual speakers bridge language gaps, given that bilinguals are driven towards CSW in cases of lack of language proficiency \cite{heredia2001bilingual}. Another interesting translation direction is translating from monolingual to CSW, which has been slightly explored in the context of CSW data augmentation \cite{hamed2023data} and ChatGPT evaluation \cite{khondaker2023gptaraeval}. 

\subsection{Automatic Speech Recognition (ASR)}
CSW presents challenges to ASR on all three fronts; data collection, modeling and evaluation. In terms of \textit{data collection}, given the costly and tedious process of building CSW speech corpora, they are usually scarce. 
CSW also usually occurs in only a portion of naturally-occurring speech, further limiting the amount of collected data. Challenges in 
\textit{modeling} include the scarcity/lack of CSW corpora and its imbalance with the vaster amounts of MSA and dialectal Arabic data that are usually used to boost performance, in addition to the mixed acoustic dynamics that can arise with using different languages \cite{HVA20,mustafa2022code}. In terms of methodologies, 
both hybrid \cite{elfahal2020framework,chowdhury2020effects,ali2021arabic,hussein2023textual} and end-to-end \cite{mubarak2021qasr,chowdhury2021towards,hamed2022arzenST,hamed2023investigating,hussein2024speech} ASR architectures have been investigated. Both architectures were compared in \citet{hamed2022investigations}, where comparable, yet complementary, performance was reported. 
Pretrained models have mostly been used for benchmarking purposes \cite{heakl2024arzen,al2024mixat,abdelali2024larabench}, where more research is needed to explore how to optimally use these models to achieve further advancements, as conducted in \citet{kulkarni2023adapting}. 
In terms of \textit{evaluation}, limitations of 
WER and CER 
have been 
discussed \cite{chowdhury2021towards,hamed2022investigations,abdallah2024leveraging}. Researchers have utilized transliteration as well as 
phonological and semantic similarities to alleviate cross-transcription issue, improving correlations with human judgments \cite{hamed2023benchmarking,kadaoui2024polywer}.

\subsection{Guidelines and Annotation Tools}
\label{sec:guidelines_annotationTools}
Guidelines have been developed for CSW Arabic data collection for several tasks, including transcriptions \cite{HVA20,hamed2024zaebuc}, ASR minimal post-editing \cite{hamed2023benchmarking}, translations \cite{hamed2022arzenST}, annotations for word-level LID in text \cite{SM16a,DGH+19} and speech \cite{chowdhury2020effects} as well as dialectness level identification \cite{badawi1973mustawayat,HRD+08}. 
With regards to useful tools, several tools have been developed for facilitating 
data collection 
\cite{BD10,VTL+14,AR15,DGH+19,elwy2024takween}.

\section{Research Gaps and Future Directions}
\label{sec:future_directions}
This survey highlights a clear need to expand research in CSW Arabic NLP to address a wider coverage of languages and tasks.  Efforts are needed across data collection, modeling, and evaluation, ultimately supporting the development of more inclusive language technologies that can effectively handle the linguistic diversity within the Arab world.   

\paragraph{Building CSW Arabic Benchmarks} 
Benchmarks are essential for standardizing evaluation, tracking progress, and guiding research. However, CSW Arabic remains underrepresented in current benchmarks. In the scope of CSW, the two main benchmarks are LinCE \cite{AKS20} and GLUECoS \cite{khanuja2020gluecos}. Arabic is only represented in LinCE, covering diglossic CSW for LID and NER tasks. In the scope of Arabic NLP, \citet{nagoudi2023dolphin} and \citet{abdelali2024larabench} offer benchmarks that cover a variety of dialects and tasks, however, CSW evaluations only cover ASR and MT tasks. The existing benchmarks are insufficient to fully reflect the current performance of models, highlighting the need for broader benchmarks that encompass a wider range of language pairs and tasks to better guide future research. 


\paragraph{Understanding and Improving the Capabilities of Pretrained Models for Arabic} 
As large language models (LLMs) and large speech models gain popularity, understanding their capabilities for CSW Arabic is crucial. Researchers evaluated pretrained models for CSW languages \cite{winata2021multilingual} and Arabic NLP \cite{khondaker2023gptaraeval,nagoudi2023dolphin,abdelali2024larabench,heakl2024arzen,al2024mixat}. However, CSW Arabic is only minimally addressed, 
where the limited evaluations reveal varying performance across tasks and language pairs. 
In ASR, word error rates reported on CSW speech corpora range from 28\% to 54\% \cite{abdelali2024larabench,heakl2024arzen,kadaoui2024polywer,lachemat2024cafe}. 
In MT, LLMs' performance varies across different corpora having different language pairs and translation directions, with reported BLEU scores ranging from 38 to 87 \cite{khondaker2023gptaraeval,heakl2024arzen}. Despite these models surpassing state-of-the-art models \cite{abdelali2024larabench,heakl2024arzen}, there remains substantial room for improvement. 
Additionally, the ability of LLMs to generate CSW Arabic text has not yet been thoroughly evaluated. Such an evaluation was conducted for South East Asian languages where varying performance was reported across languages \cite{yong2023prompting}. For Arabic, \citet{khondaker2023gptaraeval} prompted ChatGPT to convert English sentences to CSW Arabic-English. While the generated translations were found to be fluent and faithful to the original meaning, human annotators gave low scores for its CSW ability, defined as `how accurately the translated text includes code-switching'. We highlight this as another area worthy of investigation. 

\paragraph{Building CSW User-facing Applications} 
As highlighted in \citet{dougruoz2021survey}, there is a need for developing applications that interact with users in CSW language. 
This is an important capability of NLP systems in order to build user-friendly and human-like language technologies. 
Such applications are still lacking for Arabic, as in other languages. One important aspect to consider here is the dynamic behaviour of CSW, where it is affected by multiple factors. 
In the Arab region, CSW behaviour is found to be affected by external factors (e.g. topic, communication channel, and participants' roles and relationships) as well as user-related factors, including demographics (e.g. age and education) and personality (e.g. extraversion and neuroticism traits) \cite{post2015impact,alaiyed2018diglossic,aljasir2020arabic,hamed2021predicting,hamed2022code}. Accordingly, applications such as chatbots should be able to identify the appropriate language settings across situations, ranging from monolingual MSA and DA to diglossic and bilingual CSW. Given that the factors affecting CSW may differ across countries, it is important to tackle this problem from the perspective of the Arabic language and culture. 

\paragraph{Generating Personalized CSW Text}
The importance of accommodating the different linguistic styles of users in user-facing applications was emphasized in \citet{bawa2020multilingual}, where users' preferences regarding CSW chatbots were affected by their attitudes and enthusiasm towards CSW. 
However, despite CSW text generation receiving considerable attention across languages, 
research exploring the personalization aspect remains limited \cite{sengupta2023persona,mondal2022cocoa}. In the scope of Arabic, CSW data augmentation has been explored for improving performance on NER, MT, ASR, and speech translation tasks \cite{sabty2021data,hamed2023investigating,hamed2023data,hussein2023textual,hussein2024speech}, however, we still lack research on the personalization front. A major bottleneck in this pursuit is the lack of diverse datasets annotated with users' meta-data, involving diverse users' sociological and psychological profiles, as well as diverse external settings. 
Therefore, we encourage researchers to include users' meta-data as part of data collection. Also, to facilitate large-scale projects, unified guidelines are needed for collecting users' meta-data. These guidelines could benefit from the vast amount of
research investigating the factors influencing CSW in the Arab world \cite{bentahila1983motivations,albirini2011sociolinguistic,hafez2015factors,post2015impact,ali2018investigation,EHA+20,alsamhan2022codeswitching}.

\paragraph{Exploring the Effectiveness of User-adaptive NLP Models}
Previous research, focused on other languages, has demonstrated benefits of adapting NLP systems to users' CSW behaviour. In \citet{vu2013investigation}, it was shown that clustering speakers based on their CSW attitude and adapting the language model accordingly can enhance language modeling and ASR performance. 
In \citet{RSB18}, features extracted from acoustics were used to distinguish between different CSW styles, where style-specific language models showed reduction in perplexity. Given the limited but promising studies, it would be interesting to extend this line of research to improve CSW Arabic NLP models.

\paragraph{CSW in Medical and Educational Domains}
CSW serves various purposes in medical and educational domains, as investigated by a number of studies in healthcare and educational institutes in the Arab world \cite{almathkuri2016investigating,alkhudair2019professors,alhamami2020switching,alkhlaifat2020code,yang2021sociolinguistic,zaghlool2023saudi,dooly2024code}. 
In the medical domain, CSW was reported to facilitate bridging lexical gaps, addressing 
sensitive topics, maintaining interpersonal relationships, and signaling power and hierarchical dynamics. It was reported, on the other hand, that the misuse of CSW may evoke negative feelings, including suspicion among patients or concealment of information and disrespect among colleagues. In the educational domain, reported CSW functions include facilitating language acquisition, supporting better and faster comprehension, increasing
student-teacher interaction, and enlivening the class atmosphere. 
It is to be noted though that CSW behaviour is highly dynamic in these domains, affected by factors such as educational instruction language, expatriate populations (being high in some Arab countries such as the UAE), disparities in education levels and language proficiency across urban and rural cities, and disparities in socio-economic backgrounds across private and public sectors. 
All these functions and nuances of CSW need to be taken into account when developing language technologies to ensure they align with the functional and cultural needs of users. 

\paragraph{Evaluating Naturalness of Generated CSW data} As mentioned in \citet{winata2023decades}, automatically evaluating the quality of generated CSW data is an area that is still understudied. 
The task has only been tackled by \citet{kodali2024human} and \citet{kuwanto2024linguistics} 
for non-Arabic language pairs. 
In the scope of Arabic, the complexity of naturalness evaluation for CSW Arabic-English was demonstrated in \citet{hamed2023data}, where human inter-annotator agreement only reached fair agreement on pairwise Cohen Kappa. 
This is due to CSW being a speaker-dependent behaviour \cite{vu2013investigation}, where annotators' judgments may be biased towards their own CSW styles and those surrounding them. In the scope of Arabic bilingual CSW, human annotations have been collected for naturalness evaluation  \cite{hussein2023textual,hamed2023investigating,hamed2023data}, however, only involving a small number of annotators. In future work, given the subjectivity of the task, we encourage diversifying the pool of annotators to 
minimize cultural bias, in line with recommendations from \citet{hershcovich2022challenges}.

\paragraph{Handling Morphological CSW} Morphological CSW adds further complexity to NLP tasks, where performance reductions are demonstrated in MT and ASR tasks for this type of CSW \cite{gaser2023exploring,hamed2022investigations}. Given that morphological CSW typically occurs in a subset of the corpus samples, we recommend that researchers report evaluation results on it separately to fully assess the CSW capabilities of the models. In the context of CSW text generation, producing morphological CSW is challenging, as it cannot be generated through rule-based approaches. For example, in Egyptian Arabic, the suffix \<ات> \textit{At} `feminine plural form' can be attached to `event' (which has a masculine gender in Arabic), while cannot be attached to `school' (which has a feminine gender in Arabic). This limits the applicability of certain text generation methods, such as lexical replacements. Back-translation, on the other hand, was found to be capable of generating sentences with correct morphological CSW \cite{hamed2023data}. With the growing interest in CSW Arabic text generation, morphological CSW deserves more attention. Broader studies are needed to capture the diverse range of patterns varying between dialects, individuals, and domains (text versus speech), where linguistic studies on morphological CSW patterns can providing valuable support for this research direction \cite{farid2019case,kniaz2021embedded}.

\paragraph{Assessing Evaluation Metrics}
Besides CSW introducing challenges in modeling, it also poses challenges in evaluation. This has been discussed in ASR for Arabic bilingual CSW \cite{hamed2023benchmarking}, where the limitations of metrics relying on string-based matching are highlighted in the case of unstandardized orthography and cross-transcription (where words are transcribed and recognized in different scripts). The authors show that word and character error rates, despite being the widely-used metrics in ASR, are not adequate under these conditions. 
We still lack similar research assessing evaluation metrics for other tasks within CSW Arabic NLP. One foreseeable issue in bilingual CSW is the limitation of string-based matching metrics used in language generation tasks, such as transliteration and summarization, where models should not be penalized for different script choices in case of borrowed words.

\paragraph{CSW Privacy and Ethical Considerations}
There are important privacy and ethical concerns that need to be addressed when collecting CSW corpora. Given that the phenomenon occurs frequently in informal communication, primary sources for obtaining CSW data include speech, social media, and chat messages. This highlights the need for proper anonymization and privacy-preserving practices 
to protect users' identities. While researchers have worked on data anonymization in text \cite{zhou2008brief,beigi2020survey,sotolavr2021towards} and speaker anonymization in speech \cite{meyer2022speaker,meyer2023prosody,meyer2023anonymizing,meyer2024probing} for monolingual data, there is a notable gap in addressing CSW contexts. We identify this as an interesting direction, where challenges introduced by CSW would need to be tackled, such as current limitations of text-to-speech systems in handling CSW with Arabic dialects. Furthermore, from a data collection stand-point, we advise researchers to obtain informed consent from participants on the public release of collected data, as a high percentage of non-public CSW data has been highlighted in \citet{winata2023decades}, which hinders research advancements in the field. Finally, we 
emphasize the importance of developing corpora covering a wide range of languages and dialects, including underrepresented CSW language setups, to avoid marginalization of minority languages, such as Saidi Arabic (upper Egyptian dialect), Fellahi Arabic (Levantine peasant dialects), Amazigh languages, Kurdish, and Shehri/Jibbali (a modern South Arabian language). 


\paragraph{Exploring Unexplored Tasks} As shown in Table~\ref{table:nlpTasks_coverage}, only half of the NLP tasks identified in Table~\ref{table:annotation_process} for annotation 
have been explored empirically. This leaves us with a significant number of unexplored tasks, including CSW point prediction \cite{SL08}, natural language entailment, natural language understanding, question answering, humor detection and generation, sarcasm detection, semantic parsing, dialectness level estimation, summarization, speech synthesis, and text-to-speech. Moreover, among the tasks that have been covered, half are addressed in only 1-2 papers. In this large pool of understudied tasks, while word-level classification tasks are important, we recommend prioritizing high-level tasks such as question answering, text-to-speech, and speech translation, as these are essential for enhancing user-facing applications that require effective handling of CSW.

\paragraph{Improving State-of-the-art CSW Arabic Models}
We highlight that even for the well-investigated NLP tasks, there is still room for improvement. 
For word-level LID, the performance of models reached an F1 score of 91.9 for diglossic CSW \cite{ASE+19} and 95.0 
for bilingual CSW \cite{shehadi2022identifying}. As for NER, the models reached an F1 score of 85.2 
for diglossic CSW \cite{winata2021multilingual} and 79.4 for bilingual CSW \cite{sabty2020contextual}. For MT, a BLEU score of 87.2 \cite{heakl2024arzen}
is reported when translating from CSW sentences to Arabic. BLEU scores ranging from 
23.1 to 53.6 
are reported when translating to the foreign language across different CSW language pairs. For ASR, 
WER in the range of 24.8-53.8\% is reported across different CSW Arabic-foreign speech corpora. For ST, BLEU scores of 31.1 and 17.0 are reported, translating to Arabic and the foreign language, respectively \cite{hamed2022arzenST,hamed2023investigating}. While comprehensive benchmarks are required, 
as previously discussed, to assess the current state-of-art, we acknowledge that further work is needed to improve the CSW capabilities of existing models, especially in the areas of text generation and speech processing.

\section{Summary and Outlook}
In this paper, we review current efforts in code-switched Arabic NLP. Arabic code-switching presents a wide range of challenges for the NLP community, 
that are compounded by the inherent complexities of Arabic itself, including diglossia, romanization, and unstandardized orthographies in dialects. We discuss existing literature, highlight research gaps and challenges, and identify directions for future work. We hope this paper guides researchers to further advance this research area, with the goal of developing language technologies that meet the linguistic needs of the Arab world. 

\section*{Limitations}
We acknowledge that our survey is limited by a number of factors. First, the scope is confined to available literature, which may exclude emerging research and recent advancements. Second, the survey primarily focuses on major Arabic dialects and may not fully represent less-studied varieties or minority languages in the Arab World. 
Finally, the diverse linguistic and cultural contexts across the Arab world can influence code-switching patterns, which may not be uniformly covered. 

\section*{Ethics Statement}
This paper did not involve human annotation collection or the creation of new datasets. We acknowledge that research in code-switched language processing, like many NLP tasks, could lead to the development of tools that might be misused for user profiling or generating non-standard language. Our focus is on advancing nuanced understanding of language use in real-world contexts to enhance applications such as speech recognition, machine translation, and text rewriting, while adhering to ethical guidelines and mitigating potential misuse.


\bibliography{custom_short}

\appendix

\section{Language and Dialect Codes}
\label{sec:appendix_lang_codes}
We outline the codes we use for languages and dialects in Table~\ref{table:lang_dialect_codes}.

\section{Overview on Empirical Papers}
\label{sec:appendix_empirical_papers}
In Tables \ref{table:list_empirical_papers_1}-\ref{table:list_empirical_papers_4}, we provide a list of the empirical papers, specifying the covered language pairs, for text-based and speech-based NLP tasks.

\section{Overview on Resource Papers}
\label{sec:appendix_resource_papers}
In Tables \ref{table:list_resource_papers_1}-\ref{table:list_resource_papers_3}, we provide a list of the resource papers. For each text-based and speech-based NLP task, we specify the language pairs covered by the collected resources presented in the papers.

\begin{table}[ht!]
\centering
\begin{tabular}{p{4cm}l}\hline
\textbf{Language/Dialect} & \textbf{Code} \\\hline
Arabic          & ARA\\
Dialectal Arabic          & DA\\
Modern Standard Arabic          & MSA\\
Algerian Arabic       & ALG\\
Egyptian Arabic       & EGY\\
Emirati Arabic       & EMI\\
Iraqi Arabic          & IRQ\\
Jordanian Arabic      & JOR\\
Lebanese Arabic       & LEB\\
Levantine Arabic      & LEV\\
Libyian Arabic        & LIB\\
Moroccan Arabic      & MOR\\
North African Arabic  & NOR\\
Palestinian Arabic    & PAL\\
Saudi Arabian Arabic         & SAU\\
Sudanese Arabic       & SUD\\
Tunisian Arabic       & TUN\\
\hline
Foreign Language        & FOR\\
Aramaic         & ARC\\
Berber          & BER\\
Dutch           & NLD\\
English         & ENG\\
French          & FRA\\
German          & DEU \\
Hebrew          & HEB\\
Indonesian      & IND\\
Judeo Arabic    & JRB\\
Spanish         & SPA\\
\hline  
\end{tabular}
\caption{The list of codes we use for languages and dialects throughout the paper.}
\label{table:lang_dialect_codes}
\end{table}

\newpage

\begin{table*}[t]
\centering
\setlength{\tabcolsep}{3pt}
\resizebox{\textwidth}{!}{
\begin{tabular}{llllllp{5cm}} \hline
& \begin{tabular}[c]{@{}c@{}}DA-\\ Foreign\end{tabular} & 

\begin{tabular}[c]{@{}c@{}}MSA-\\ Foreign\end{tabular}
 & 
 \begin{tabular}[c]{@{}c@{}}ARA-\\ Foreign\end{tabular}
 & 
 \begin{tabular}[c]{@{}c@{}}MSA-\\ DA\end{tabular}
 & 
 \begin{tabular}[c]{@{}l@{}}MSA-DA-\\ Foreign\end{tabular}
 & \multicolumn{1}{c}{Language Details}\\\hline
\multicolumn{7}{c}{Word-level Language Identification}\\\hline
\cite{EAH+14} &  \checkmark  &    &   &  &   & EGY-ENG \\
\cite{VTL+14} & \checkmark   &    &   &   &   & MOR-ENG-FRA \\
\cite{BDW+15} & \checkmark &  &   &   &   & JRB-HEB \\
\cite{aghzal2021distributional} &  \checkmark  &    &   &  &   & MOR-ENG-FRA \\
\cite{mitelman2024code} & \checkmark &  &   &   &   &JRB-HEB/ARC \\
\cite{kalkman2024detecting} &  \checkmark  &    &   &  &   & MOR-NLD \\
\cite{darwish2014arabizi} &   &    &  \checkmark &   &   & ARA-ENG \\
\cite{sabty2021language} &   &    &  \checkmark &   &   & ARA-ENG \\
\cite{shehadi2022identifying} &   &    &  \checkmark &   &   & ARA-ENG-FRA \\
\cite{ED12} &    &    &   & \checkmark  &   & MSA-EGY, MSA-LEV \\
\cite{EAD13} &    &    &   & \checkmark  &   & MSA-EGY \\
\cite{elfardy2014aida} &    &    &   & \checkmark &   & MSA-EGY \\
\cite{SBM+14} &    &    &   & \checkmark  &   & MSA-EGY \\
\cite{KBG+14} &    &    &   & \checkmark &   & MSA-EGY \\
\cite{LAL+14} &    &    &   & \checkmark &   & MSA-EGY \\
\cite{JB14} &    &    &   & \checkmark &   & MSA-EGY \\
\cite{AED15} &    &    &   & \checkmark  &   & MSA-EGY \\
\cite{AD16} &    &    &   & \checkmark  &   & MSA-EGY \\
\cite{MAG+16} &    &    &   & \checkmark  &   & MSA-EGY \\
\cite{SMA+16} &    &    &   & \checkmark &   & MSA-EGY \\
\cite{Shr16} &    &    &   & \checkmark &   & MSA-EGY \\
\cite{jaech2016neural} &    &    &   & \checkmark &   & MSA-EGY \\
\cite{ASE+19} &    &    &   & \checkmark &   & MSA-EGY \\
\cite{AKS20} &    &    &   & \checkmark &   & MSA-EGY \\
\cite{AB16} & \checkmark  &    &   & \checkmark &   & MSA-EGY/LEV/GLF, DA-ENG\\
\cite{SSF+18} & \checkmark  &    &   & \checkmark &   & MSA, ALG/TUN/MOR/EGY, ENG/FRA
\\
\cite{SM16a} &    &   &   &   &  \checkmark & MSA-MOR-FRA/SPA/BER \\
\cite{AD17} &    &   &   &   &  \checkmark & MSA-ALG-ENG-FRA-BER \\
\cite{ADB+18} &    &   &   &   &  \checkmark & MSA-ALG-ENG-FRA-BER \\
\cite{ABD+18} &    &   &   &   &  \checkmark & MSA-ALG-ENG-FRA-BER \\
\cite{adouane2020multi} &    &   &   &   &  \checkmark & MSA-ALG-ENG-FRA-BER \\
\cite{TTA+20} &    &    &   &   &  \checkmark & MSA-EGY-SAU-ENG \\
\hline
\multicolumn{7}{c}{Named Entity Recognition}\\\hline
\cite{SEA19} & \checkmark & \checkmark  &   &  &  & MSA-ENG, EGY-ENG \\
\cite{SSE+19} & \checkmark &  \checkmark &   &  &  & MSA-ENG, EGY-ENG \\
\cite{sabty2020contextual}  & \checkmark &  \checkmark &  &  &  & MSA-ENG, EGY-ENG \\
\cite{sabty2021data} & \checkmark & \checkmark &  &  &  & MSA-ENG, EGY-ENG \\
\cite{AAS+18b} &    &   &   & \checkmark &  & MSA-EGY \\
\cite{WCK18} &    &   &   & \checkmark &  & MSA-EGY \\
\cite{JKD+18} &    &   &   & \checkmark &  & MSA-EGY \\
\cite{GCB18} &    &   &   & \checkmark &  & MSA-EGY \\
\cite{janke2018university} &    &   &   & \checkmark &  & MSA-EGY \\
\cite{ASM17} &    &   &   & \checkmark &  & MSA-EGY \\
\cite{AKS20} &    &   &   & \checkmark &  & MSA-EGY \\
\cite{chopra2021switch} &    &   &   & \checkmark &  & MSA-EGY \\
\cite{winata2021multilingual} &    &   &   & \checkmark &  & MSA-EGY \\
\cite{eskander2022towards} &    &   &   & \checkmark &  & MSA-EGY \\

\cite{adouane2020multi}  &  &   &  &  & \checkmark & MSA-ALG-ENG-FRA-BER \\
\hline
\end{tabular}
}
\caption{List of empirical papers for the stated text-based tasks.}
\label{table:list_empirical_papers_1}
\end{table*}

\begin{table*}[t]
\centering
\setlength{\tabcolsep}{3pt}
\resizebox{\textwidth}{!}{
\begin{tabular}{llllllp{5cm}} \hline
& \begin{tabular}[c]{@{}c@{}}DA-\\ Foreign\end{tabular} & 

\begin{tabular}[c]{@{}c@{}}MSA-\\ Foreign\end{tabular}
 & 
 \begin{tabular}[c]{@{}c@{}}ARA-\\ Foreign\end{tabular}
 & 
 \begin{tabular}[c]{@{}c@{}}MSA-\\ DA\end{tabular}
 & 
 \begin{tabular}[c]{@{}l@{}}MSA-DA-\\ Foreign\end{tabular}
 & \multicolumn{1}{c}{Language Details}\\\hline
\multicolumn{7}{c}{Machine Translation}\\\hline
\cite{hamed2022arzenST}  & \checkmark &  &  &  &  & EGY-ENG \\
\cite{hamed2023data}  & \checkmark &  &  &  &  & EGY-ENG \\
\cite{gaser2023exploring}  & \checkmark &  &  &  &  & EGY-ENG \\
\cite{hamed2023investigating}  & \checkmark &  &  &  &  & EGY-ENG \\
\cite{khondaker2023gptaraeval}  & \checkmark &  &  &  &  & ALG-FRA, JOR-ENG\\
\cite{nagoudi2023dolphin}  & \checkmark &  &  &  &  & ALG/MOR/TUN-FRA, \newline EGY/JOR/PAL-ENG \\
\cite{khondaker2024benchmarking}  & \checkmark &  &  &  &  & ALG/MOR/TUN-FRA, \newline EGY/JOR/PAL-ENG \\
\cite{heakl2024arzen}  & \checkmark &  &  &  &  & EGY-ENG \\
\cite{MLJ+19}  &  & \checkmark &  &  &  & MSA-ENG \\
\cite{elmadany2022arat5}  & \checkmark & \checkmark &  &  &  & ALG-FRA, JOR-ENG, \newline (synthetic) MSA-ENG/FRA \\
\cite{saadany2020great}  &  &  &  & \checkmark &  & MSA-DA (DA mostly EGY) \\
\cite{elmadany2021investigating}  &  &  &  & \checkmark &  & MSA-EGY \\
\cite{chen2021calcs}  &  &  &  & \checkmark &  & MSA-EGY \\
\hline
\multicolumn{7}{c}{Sentiment Analysis}\\\hline
\cite{TFA+19} & \checkmark &  &  &  &  & LEB-ENG\\
\cite{touileb2021interplay} &  \checkmark  &   &   &   &  & ALG-FRA \\
\cite{boudad2023multilingual} & \checkmark &  &  &  &  & MOR-FOR\\
\cite{sherif2025sentiment} & \checkmark &   &   &   &  & EGY-ENG \\
\cite{sabty2020contextual}  & \checkmark & \checkmark  &  &  &  & MSA-ENG, EGY-ENG \\
\cite{almasah2023code}  &  &   & \checkmark &  &  & ARA-FOR \\
\cite{JAS19} &  &  &  &  & \checkmark & MSA-TUN-ENG-FRA\\
\cite{mataoui2016proposed} &  &  &  &  & \checkmark & MSA-ALG-FRA\\
\cite{adouane2020identifying} &  &  &  &  & \checkmark & MSA-ALG-FRA-BER\\
\cite{adouane2020multi} &    &   &   &   &  \checkmark & MSA-ALG-ENG-FRA-BER \\
\hline
  \multicolumn{7}{c}{Part-of-Speech Tagging}\\\hline
\cite{MSS20} & \checkmark &  &  &  &  & ALG-FRA \\
\cite{SEF+20} &  \checkmark &    &   &   &   & ALG-FRA \\
\cite{touileb2021interplay} & \checkmark &  &  &  &  & ALG-FRA \\
\cite{cheragui2023exploring} & \checkmark &  &  &  &  & ALG-FOR \\
\cite{AMD+16} &  &  &  & \checkmark &  & MSA-EGY \\
\cite{AD19} &  &  &  & \checkmark &  & MSA-EGY/LEV \\
\hline
\multicolumn{7}{c}{Sentence-level Language Identification}\\\hline
\cite{shehadi2022identifying} &  &  & \checkmark &  &  & ARA-ENG-FRA \\
\cite{AED15} &    &    &   & \checkmark  &   & MSA-EGY \\
\cite{burchell2024code} &    &    &   & \checkmark  &   & MSA-EGY \\
\cite{ERA18} &  &  &  & \checkmark &  & MSA-GLF/LEV/TUN/EGY \\
\cite{Alt20} &  &  &  & \checkmark &  & MSA-EGY/GLF/IRQ \newline MSA-LEV/Maghrebi \\
\hline
\end{tabular}
}
\caption{List of empirical papers for the stated text-based tasks.}
\label{table:list_empirical_papers_2}
\end{table*}

\begin{table*}[t]
\centering
\setlength{\tabcolsep}{3pt}
\resizebox{\textwidth}{!}{
\begin{tabular}{llllllp{5cm}} \hline
& \begin{tabular}[c]{@{}c@{}}DA-\\ Foreign\end{tabular} & 

\begin{tabular}[c]{@{}c@{}}MSA-\\ Foreign\end{tabular}
 & 
 \begin{tabular}[c]{@{}c@{}}ARA-\\ Foreign\end{tabular}
 & 
 \begin{tabular}[c]{@{}c@{}}MSA-\\ DA\end{tabular}
 & 
 \begin{tabular}[c]{@{}l@{}}MSA-DA-\\ Foreign\end{tabular}
 & \multicolumn{1}{c}{Language Details}\\\hline
\multicolumn{7}{c}{Transliteration}\\\hline
\cite{AEH+14}  & \checkmark &  &  &  &  & EGY-ENG\\
\cite{EAH+14}  & \checkmark &  &  &  &  & EGY-ENG\\
\cite{shazal2020unified}  & \checkmark &  &  &  &  & EGY-ENG\\
\cite{mitelman2024code} & \checkmark  &  & &  &  & JRB-HEB/ARC\\
\cite{darwish2014arabizi}  &  &  & \checkmark &  &  & ARA-ENG\\
\hline
 \multicolumn{7}{c}{Language Modeling}\\
\multicolumn{7}{c}{(Excluding language modeling conducted as part of ASR efforts)}\\\hline
\cite{HZE+19} & \checkmark &  &  &  &  & EGY-ENG\\
\cite{HEA17} &  & \checkmark &  &  &  & MSA-ENG\\
\cite{HEA18b} &  & \checkmark &  &  &  & MSA-ENG\\
\cite{lan2020empirical} &  &  & \checkmark &  &  & ARA-ENG\\

\hline
\multicolumn{7}{c}{Abusive Language Detection}\\\hline
\cite{guellil2021sexism} & \checkmark &  &  &  &  & ALG-ENG-FRA \\
\cite{alhazmi2024code} &  &  & \checkmark &  &  & ARA-FOR \\
\hline
 \multicolumn{7}{c}{Sentence-level Micro-Dialect Identification}\\\hline
\cite{Aba18} & \checkmark &  &  &  &  & ALG-FRA \\
\cite{abdul2020toward} &  &  & \checkmark &  &  & ARA-FOR \\
\hline
\multicolumn{7}{c}{Spelling Correction and Text Normalization}\\\hline
\cite{ABD19} &  &  &  &  & \checkmark & MSA-ALG-ENG-FRA-BER \\
\cite{adouane2020multi} &  &  &  &  & \checkmark & MSA-ALG-ENG-FRA-BER \\
\hline
\multicolumn{7}{c}{Dependency Parsing}\\\hline
\cite{MSS20} & \checkmark &  &  &  &  & ALG-FRA \\
\cite{SEF+20} &  \checkmark &    &   &   &   & ALG-FRA \\
\hline
\multicolumn{7}{c}{Tokenization}\\\hline
\cite{gaser2023exploring} & \checkmark &  &  &  &  & EGY-ENG \\
\hline
\multicolumn{7}{c}{Fake News Detection}\\\hline
\cite{abdelouahab2024detecting} &  &  &  &  & \checkmark & MSA-TUN/ALG-ENG-FRA\\
\hline
\multicolumn{7}{c}{Word Analogy}\\\hline
\cite{aghzal2021distributional} & \checkmark &  &  &  &  & MOR-ENG-FRA \\\hline
\multicolumn{7}{c}{Topic Modeling}\\\hline
\cite{touileb2021interplay} &  \checkmark  &   &   &   &  & ALG-FRA \\\hline
\multicolumn{7}{c}{Question Answering}\\\hline
\cite{sabty2020contextual}  & \checkmark &  \checkmark &  &  &  & MSA-ENG, EGY-ENG \\\hline
\end{tabular}
}
\caption{List of empirical papers for the stated text-based tasks.}
\label{table:list_empirical_papers_3}
\end{table*}

\begin{table*}[t]
\centering
\setlength{\tabcolsep}{3pt}
\resizebox{\textwidth}{!}{
\begin{tabular}{llllllp{5cm}} \hline
& \begin{tabular}[c]{@{}c@{}}DA-\\ Foreign\end{tabular} & 

\begin{tabular}[c]{@{}c@{}}MSA-\\ Foreign\end{tabular}
 & 
 \begin{tabular}[c]{@{}c@{}}ARA-\\ Foreign\end{tabular}
 & 
 \begin{tabular}[c]{@{}c@{}}MSA-\\ DA\end{tabular}
 & 
 \begin{tabular}[c]{@{}l@{}}MSA-DA-\\ Foreign\end{tabular}
 & \multicolumn{1}{c}{Language Details}\\\hline
 \multicolumn{7}{c}{Automatic Speech Recognition}\\\hline
 \cite{Elf19} &  \checkmark &  &  &  &  & SUD-ENG \\
 \cite{elfahal2020framework} &  \checkmark &  &  &  &  & SUD-ENG \\
 \cite{lounnas2021towards} &  \checkmark &  &  &  &  & ALG-FRA \\
 \cite{hamed2022investigations} &  \checkmark &  &  &  &  & EGY-ENG \\
 \cite{hamed2022arzenST} &  \checkmark &  &  &  &  & EGY-ENG \\
 \cite{hamed2023investigating} &  \checkmark &  &  &  &  & EGY-ENG \\
 \cite{abdallah2024leveraging} &  \checkmark &  &  &  &  & TUN-ENG-FRA \\
 \cite{heakl2024arzen} &  \checkmark &  &  &  &  & EGY-ENG \\
 \cite{al2024mixat} &  \checkmark &  &  &  &  & EMI-ENG \\
 \cite{al2024enhancing} &  \checkmark &  &  &  &  & EMI-ENG \\
 \cite{lachemat2024cafe} & \checkmark &  &  &  &  & ALG-ENG-FRA\\
 \cite{BL19} & &  & \checkmark &  &  & ARA-IND\\
 \cite{ugan2022language} & &  & \checkmark &  &  & ARA-DEU\\
  \cite{ali2021arabic} & &  & \checkmark &  &  & ARA-ENG/FRA\\
 \cite{kulkarni2023adapting} & &  & \checkmark &  &  & ARA-ENG\\
 \cite{hussein2023textual} & &  & \checkmark &  &  & ARA-ENG\\
 \cite{hussein2024speech} & &  & \checkmark &  &  & ARA-ENG\\
 \cite{kadaoui2024polywer} & &  & \checkmark &  &  & ARA-ENG\\
 \cite{chowdhury2020effects} & &  &  & \checkmark &  & MSA-EGY\\
 \cite{chowdhury2021towards} & & & \checkmark & \checkmark &  & MSA-EGY, ARA-ENG/FRA\\
 \cite{mubarak2021qasr} & & & \checkmark & \checkmark &  & MSA-DA, ARA-ENG/FRA\\
 \cite{abdelali2024larabench} & & & \checkmark & \checkmark &  & ARA-ENG/FRA, MSA-DA \newline(DA: EGY, GLF, LEV, NOR) \\
 \hline
  \multicolumn{7}{c}{Speech Translation}\\\hline
 \cite{hamed2023investigating} & \checkmark &  &  &  &  & EGY-ENG \\
 \cite{hamed2022arzenST} &  \checkmark &  &  &  &  & EGY-ENG \\
 \hline
 \multicolumn{7}{c}{Word-level Language Identification}\\\hline
 \cite{chowdhury2020effects} & &  &  & \checkmark &  & MSA-EGY \\
 \hline
 \multicolumn{7}{c}{Sentence Boundary Detection}\\\hline
 \cite{ZIE+16} & &  &  &  & \checkmark & MSA-TUN-FRA \\
 \hline
\end{tabular}
}
\caption{List of empirical papers for the stated speech-based tasks.}
\label{table:list_empirical_papers_4}
\end{table*}

\begin{table*}[t]
\centering
\setlength{\tabcolsep}{3pt}
\resizebox{\textwidth}{!}{
\begin{tabular}{llllllp{5cm}} \hline
& \begin{tabular}[c]{@{}c@{}}DA-\\ Foreign\end{tabular} & 

\begin{tabular}[c]{@{}c@{}}MSA-\\ Foreign\end{tabular}
 & 
 \begin{tabular}[c]{@{}c@{}}ARA-\\ Foreign\end{tabular}
 & 
 \begin{tabular}[c]{@{}c@{}}MSA-\\ DA\end{tabular}
 & 
 \begin{tabular}[c]{@{}l@{}}MSA-DA-\\ Foreign\end{tabular}
 & \multicolumn{1}{c}{Language Details}\\\hline
\multicolumn{7}{c}{Word-level Language Identification}\\\hline
\cite{VTL+14} &  \checkmark &    &   &   &   & MOR-ENG-FRA \\
\cite{CRS+14} &  \checkmark &    &   &   &   & ALG-FRA \\
\cite{AAL17} &  \checkmark &    &   &   &   & ALG-FRA \\
\cite{AAM18} &  \checkmark &    &   &   &   & ALG-FRA \\
\cite{Aba19} &  \checkmark &    &   &   &   & ALG-ENG/FRA \\
\cite{SEF+20} &  \checkmark &    &   &   &   & ALG-FRA \\
\cite{kalkman2024detecting}  &  \checkmark &    &   &   &   & MOR-NLD \\
\cite{darwish2014arabizi} &    &    &  \checkmark  &   &   & ARA-ENG\\
\cite{sabty2021language} &    &    &  \checkmark  &   &   & ARA-ENG \\
\cite{shehadi2022identifying} &    &    &  \checkmark  &   &   & ARA-ENG-FRA \\

\cite{HRD+08} &    &    &   & \checkmark  &   & MSA-DA \\
\cite{ED12} &   &    &   &  \checkmark &   & MSA-EGY, MSA-LEV \\
\cite{SBM+14} &   &    &   &  \checkmark &   & MSA-EGY \\
\cite{MAG+16} &    &    &   & \checkmark  &   & MSA-EGY \\
\cite{chowdhury2020effects} &    &    &    &  \checkmark &   & MSA-EGY \\

\cite{samih2016arabic} &   &    &   &   &  \checkmark & MSA-MOR-ENG/FRA/SPA/BER \\
\cite{SM16a} &   &    &   &   &  \checkmark & MSA-MOR-FRA/SPA/BER \\
\cite{AS17} &   &    &   &   &  \checkmark & MSA-ALG-ENG-FRA\\
\cite{AD17} &   &    &   &   &  \checkmark & MSA-ALG-ENG-FRA-BER \\
\cite{DGH+19} &   &    &   &   & \checkmark  & MSA-EGY-FOR\\
\cite{TTA+20} &   &    &   &   &  \checkmark & MSA-EGY-SAU-ENG\\
\cite{hajbi2022natural} &   &    &   &   &  \checkmark & MSA-MOR-ENG/FRA/SPA\\
\hline

\multicolumn{7}{c}{Machine Translation}\\\hline
\cite{BSM+14}  &\checkmark &    &   &   &   & EGY-ENG \\
\cite{Aba19} &  \checkmark &    &   &   &   & ALG-ENG/FRA \\
\cite{SEF+20} &  \checkmark &    &   &   &   & ALG-FRA \\
\cite{hamed2022arzenST} & \checkmark  &    &   &   &   & EGY-ENG \\
\cite{nagoudi2023dolphin} & \checkmark &  &  &  &  & ALG/MOR/TUN-FRA, EGY/JOR/PAL-ENG \\
\cite{MLJ+19} &   &  \checkmark  &   &   &   & MSA-ENG \\
\cite{elmadany2022arat5}   & \checkmark & \checkmark &  &  &  & ALG-FRA, JOR-ENG, \newline (synthetic) MSA-ENG/FRA \\
\hline

\multicolumn{7}{c}{Transliteration}\\\hline
\cite{AEH+14} &  \checkmark &    &   &   &   & EGY-ENG \\
\cite{BSM+14}  &\checkmark &    &   &   &   & EGY-ENG \\
\cite{chen2017bolt}  & \checkmark &  &  &  &  & EGY-FOR\\
\cite{Aba19} &  \checkmark &    &   &   &   & ALG-ENG/FRA \\
\cite{touileb2021interplay} & \checkmark &  &  &  &  & ALG-FRA \\
\cite{darwish2014arabizi}  &  &  & \checkmark &  &  & AR-ENG\\
\cite{alhazmi2024code} &  &  & \checkmark &  &  & ARA-FOR \\
\hline

\multicolumn{7}{c}{Sentiment Analysis}\\\hline
\cite{TFA+19} & \checkmark  &    &   &   &   & LEB-ENG\\
\cite{sherif2025sentiment} & \checkmark &   &   &   &  & EGY-ENG \\
\cite{almasah2023code}  &  &   & \checkmark &  &  & ARA-FOR \\
\cite{mataoui2016proposed} &   &    &   &   & \checkmark & MSA-ALG-FRA \\
\cite{adouane2020identifying} &  &  &  &  & \checkmark & MSA-ALG-FRA-BER\\
\cite{touileb2021interplay} & \checkmark &  &  &  &  & ALG-FRA \\
\hline
\end{tabular}
}
\caption{List of resource papers supporting the stated text-based tasks.}
\label{table:list_resource_papers_1}
\end{table*}

\begin{table*}[t]
\centering
\setlength{\tabcolsep}{3pt}
\resizebox{\textwidth}{!}{
\begin{tabular}{llllllp{5cm}} \hline
& \begin{tabular}[c]{@{}c@{}}DA-\\ Foreign\end{tabular} & 

\begin{tabular}[c]{@{}c@{}}MSA-\\ Foreign\end{tabular}
 & 
 \begin{tabular}[c]{@{}c@{}}ARA-\\ Foreign\end{tabular}
 & 
 \begin{tabular}[c]{@{}c@{}}MSA-\\ DA\end{tabular}
 & 
 \begin{tabular}[c]{@{}l@{}}MSA-DA-\\ Foreign\end{tabular}
 & \multicolumn{1}{c}{Language Details}\\\hline
 \multicolumn{7}{c}{Named Entity Recognition}\\\hline
\cite{SEA19} & \checkmark & \checkmark  &   &  &  & MSA-ENG, EGY-ENG \\
\cite{SSE+19}& \checkmark & \checkmark  &   &  &  & MSA-ENG, EGY-ENG \\
\cite{AAS+18b} &   &    &   &  \checkmark &   & MSA-EGY \\
\hline
\multicolumn{7}{c}{Sentence-level Language Identification}\\\hline
\cite{shehadi2022identifying} &    &    &  \checkmark  &   &   & ARA-ENG-FRA\\
\cite{altamimi2018btac} &   &    &   &  \checkmark &   & MSA-EGY/GLF/IRQ/LEV, \newline MSA-MOR/TUN/LIB/ALG \\
\cite{chowdhury2020effects} &    &    &    &  \checkmark &   & MSA-EGY \\
\hline

\multicolumn{7}{c}{Tokenization}\\\hline
\cite{BHA+20} &  \checkmark &    &   &   &   & EGY-ENG \\
\cite{SEF+20} &  \checkmark &    &   &   &   & ALG-FRA \\
\cite{gaser2023exploring} &  \checkmark &    &   &   &   & EGY-ENG \\
\hline

\multicolumn{7}{c}{Part-of-speech Tagging}\\\hline
\cite{BHA+20} &  \checkmark &    &   &   &   & EGY-ENG \\
\cite{SEF+20} &  \checkmark &    &   &   &   & ALG-FRA \\
\cite{DGH+19} &   &    &   &   & \checkmark  & MSA-EGY-FOR\\
\hline

\multicolumn{7}{c}{Abusive Language Detection}\\\hline
\cite{Aba19} &  \checkmark &    &   &   &   & ALG-ENG/FRA \\
\cite{guellil2021sexism} &  \checkmark &    &   &   &   & ALG-ENG/FRA \\
\cite{alhazmi2024code} &  &  & \checkmark &  &  & ARA-FOR \\
\hline

\multicolumn{7}{c}{Sentence-level Micro-Dialect Identification}\\\hline
\cite{Aba18} & \checkmark  &    &   &   &   & ALG-FRA \\
\cite{abdul2020toward} &   &   &  \checkmark  &   &   & ARA-FOR \\
\hline

\multicolumn{7}{c}{Dialectness Level Estimation}\\\hline
\cite{HRD+08} &    &    &   & \checkmark  &   & MSA-DA \\
\cite{hamed2024zaebuc} &   &    &   &   & \checkmark  & MSA-EMI/EGY-ENG \\
\hline

\multicolumn{7}{c}{Spelling Correction and Text Normalization}\\\hline
\cite{ABD19} &   &    &   &   &  \checkmark & MSA-ALG-ENG-FRA-BER\\
\hline

\multicolumn{7}{c}{Dependency Parsing}\\\hline
\cite{SEF+20} &  \checkmark &    &   &   &   & ALG-FRA \\
\hline

\multicolumn{7}{c}{Fake News Detection}\\\hline
\cite{smaili2024boutef} &   &    &   &   & \checkmark  & MSA-TUN/ALG-ENG-FRA\\
\hline

\multicolumn{7}{c}{Word Analogy}\\\hline
\cite{aghzal2021distributional} & \checkmark  &    &   &   &   & MOR-ENG-FRA\\
\hline

\multicolumn{7}{c}{Lemmzatization}\\\hline
\cite{BHA+20} &  \checkmark &    &   &   &   & EGY-ENG \\
\hline

\multicolumn{7}{c}{Emotion Detection}\\\hline
\cite{Aba19} &  \checkmark &    &   &   &   & ALG-ENG/FRA \\\hline
\multicolumn{7}{c}{Topic Modeling}\\\hline
\cite{touileb2021interplay} &  \checkmark  &   &   &   &  & ALG-FRA \\\hline
\multicolumn{7}{c}{Text Corpora (without annotations)}\\\hline
\cite{HZE+19} & \checkmark &  &  &  &  & EGY-ENG\\
\cite{aghzal2021distributional} & \checkmark  &    &   &   &   & MOR-ENG-FRA\\
\cite{HEA17} &   &  \checkmark  &   &   &   & MSA-ENG \\
\cite{ADB+18} &   &    &   &   &  \checkmark & MSA-ALG-ENG-FRA-BER \\
\hline
\end{tabular}
}
\caption{List of resource papers supporting the stated text-based tasks.}
\label{table:list_resource_papers_2}
\end{table*}

\begin{table*}[t]
\centering
\setlength{\tabcolsep}{3pt}
\resizebox{\textwidth}{!}{
\begin{tabular}{llllllp{5cm}} \hline
& \begin{tabular}[c]{@{}c@{}}DA-\\ Foreign\end{tabular} & 

\begin{tabular}[c]{@{}c@{}}MSA-\\ Foreign\end{tabular}
 & 
 \begin{tabular}[c]{@{}c@{}}ARA-\\ Foreign\end{tabular}
 & 
 \begin{tabular}[c]{@{}c@{}}MSA-\\ DA\end{tabular}
 & 
 \begin{tabular}[c]{@{}l@{}}MSA-DA-\\ Foreign\end{tabular}
 & \multicolumn{1}{c}{Language Details}\\\hline
\multicolumn{7}{c}{Automatic Speech Recognition }\\\hline
\cite{Ism15} & \checkmark  &    &   &   &   & SAU-ENG \\
\cite{AAL17} & \checkmark  &    &   &   &   & ALG-FRA \\
\cite{AAM18} & \checkmark  &    &   &   &   & ALG-FRA \\
\cite{HEA18} & \checkmark  &    &   &   &   & EGY-ENG \\
\cite{Elf19} &  \checkmark &  &  &  &  & SUD-ENG \\
\cite{HVA20} & \checkmark  &    &   &   &   & EGY-ENG \\
\cite{lounnas2021towards} &  \checkmark &  &  &  &  & ALG-FRA \\
\cite{hamed2022arzenST} & \checkmark  &    &   &   &   & EGY-ENG \\
\cite{abdallah2024leveraging} & \checkmark  &    &   &   &   & TUN-ENG-FRA \\
\cite{al2024mixat} & \checkmark  &    &   &   &   & EMI-ENG \\
\cite{lachemat2024cafe} & \checkmark  &    &   &   &   & ALG-ENG-FRA \\
\cite{ali2021arabic} & &  & \checkmark &  &  & ARA-ENG/FRA\\
\cite{ugan2022language} & &  & \checkmark &  &  & ARA-DEU\\
\cite{chowdhury2020effects} &   &    &  & \checkmark  &   & MSA-EGY \\
\cite{mubarak2021qasr} &   &    & \checkmark  & \checkmark  &   & MSA-DA, ARA-ENG/FRA \\
\cite{zribi2015spoken} &   &    &   &   &  \checkmark & MSA-TUN-FRA \\
\cite{hamed2024zaebuc} &   &    &   &   & \checkmark  & MSA-EMI/EGY-ENG \\
\hline

\multicolumn{7}{c}{Word-level Language Identification}\\\hline
\cite{MML16} &  \checkmark &    &   &   &   & ALG/TUN/MOR-FRA \\
\cite{AAL17} &  \checkmark &    &   &   &   & ALG-FRA \\
\cite{AAM18} &  \checkmark &    &   &   &   & ALG-FRA \\
\cite{chowdhury2020effects} &   &    &   & \checkmark  &   & MSA-EGY \\
\hline

\multicolumn{7}{c}{Sentence-level Language Identification}\\\hline
\cite{chowdhury2020effects} &   &    &   & \checkmark  &   & MSA-EGY \\
\hline

\multicolumn{7}{c}{Speech Translation}\\\hline
\cite{hamed2022arzenST} & \checkmark  &    &   &   &   & EGY-ENG \\

\hline

\end{tabular}
}
\caption{List of resource papers supporting the stated speech-based tasks.}
\label{table:list_resource_papers_3}
\end{table*}
\end{document}